# Normative Engineering Risk Management Systems

Peter J. Regan
Engineering-Economic Systems Department
Stanford University
Stanford, CA 94305

## Abstract

This paper describes a normative system design that incorporates diagnosis, dynamic evolution, decision making, and information gathering. A single influence diagram demonstrates the design's coherence, yet each activity is more effectively modeled and evaluated separately. Application to offshore oil platforms illustrates the design. For this application, the normative system is embedded in a real-time expert system.

## 1  INTRODUCTION

Recent years have witnessed great strides in normative model evaluation methods that can support more sophisticated normative systems. "Normative" herein refers to using probability to represent uncertainty and expected utility maximization to choose among decision alternatives. Evaluation algorithms for influence diagrams (decisions, uncertainties, and values) and belief networks (uncertainties only) are the computational foundation for a new generation of software tools (e.g., DAVID, DA Workbench, DEMOS, DPL, ERGO, HUGIN, IDEAL).

By extending the problem evaluation frontier, these software tools facilitate normative system designs that address more fully the following features:

- Large Models
- Sequential Decisions
- Asymmetric Model Structure
- Dynamic Physical System

This paper describes a normative design that separates diagnosis, dynamic evolution, decision making, and information gathering to better match problem formulation and evaluation with each activity. This design is applied to an industrial setting. The normative system is embedded in a real-time expert system to create a powerful engineering risk management system (ERMS). Combining normative and nonnormative technologies allows normative power to be used for complex uncertain reasoning tasks while relying on traditional approaches where a normative perspective offers no substantial benefit.

The paper is organized as follows. Section 2 introduces ERMS design. Sections 3 and 4 introduce the Advanced Risk Management System (ARMS) project and overview normative systems, respectively. Section 5 presents the normative system activities with illustrations from the ARMS project. Sections 6 and 7 introduce research issues and present conclusions.

## 2  ENGINEERING RISK MANAGEMENT SYSTEMS

Probabilistic risk analysis (PRA) examines low-probability high-consequence events in complex systems (Henley and Kumamoto 1981, US NRC 1975). PRA is often performed to validate a proposed engineering system design. Engineering risk management determines the response to risk for an operating industrial system (Paté-Cornell 1990). Decision analysis (Howard 1966, Howard and Matheson 1981), with its synthesis of alternatives, information, and preferences, and its focus on clarity of action, is well suited to address engineering risk management. Normative systems deliver decision analytic assistance rapidly in an operating environment.

Engineering risk management problems exhibit a characteristic profile (Figure 1). Following an initiating event, the physical system degrades. Warning signal thresholds balance false alerts and lead time for response to an abnormal situation. Delay between signal and action balances system degradation and information gathering. Alternative actions balance cost and improvement. The profile diverges to represent the effects of different actions.

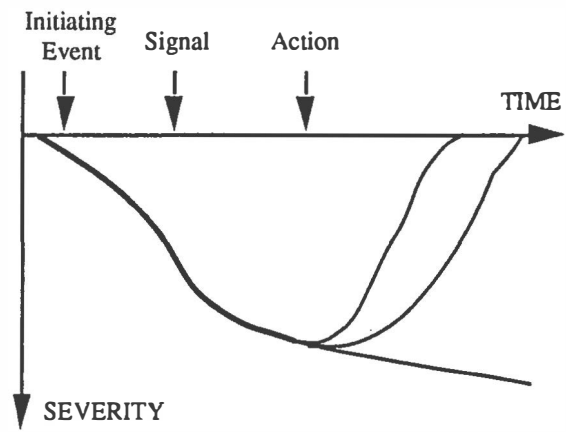

Figure 1: Engineering Risk Management Profile

The ERMS design challenge is to identify and characterize risk and to recommend a decision-making and information-gathering response. Four activities are highlighted, each phrased in terms of a simple question:

Diagnosis:            What is the risk at
                      a given time?

Dynamic Evolution:    How does the risk
                      change over time?

Decision Making:      What is the preferred action
                      given the risk and its change
                      over time?

Information Gathering: What information should be
                      obtained to refine the preferred
                      action?

Although this paper concentrates on engineering risk management, the concepts generalize to other domains (e.g., medicine, business).

Five decision analysis process steps provide a framework for this paper's contribution. Formulation creates a model that links the principal decision to value. Evaluation computes a principal decision recommendation. Appraisal identifies principal decision model uncertainties with high value of perfect information. Reformulation modifies the principal decision model to represent the cost and reliability of alternatives to gather imperfect information. Reevaluation computes an information-gathering and principal decision recommendation.

Normative diagnostic system (Heckerman et al. 1990) models are characteristic of the reformulation step. Myopic reevaluation is their basic activity. Intelligent decision systems (Holtzman 1989) interactively guide a user through the formulation, evaluation, and appraisal steps. This paper joins an IDS's decision focus with a normative diagnostic system's imperfect information focus, adds a dynamic physical system model, and introduces an algorithm to guide automatically the reformulation and reevaluation steps.

## 3 ADVANCED RISK MANAGEMENT SYSTEM PROJECT

### 3.1 OVERVIEW

The Advanced Risk Management System (ARMS) (Besse et al. 1992) project's objective is to design and develop a proof-of-concept ERMS for offshore oil platform operations.[1] The project is motivated by the 1988 Piper Alpha platform explosion in the North Sea that resulted in over 160 deaths (Cullen 1990).

---

[1] The ARMS project is funded primarily by the European Community and is coordinated by Bureau Veritas

Pumping, separation, gas compression, and transport are an offshore platform's major functions (Figure 2). A hydrocarbon mixture is pumped from the subsea reservoir to the platform where it is separated. The oil is pumped directly to shore. The gas is compressed and pumped to shore. Water is treated and discharged to the sea.

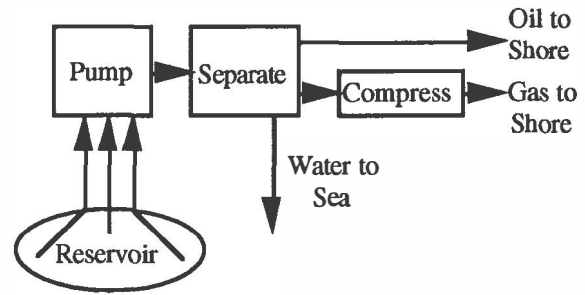

Figure 2: Offshore Oil Platform Flow Diagram

Compared to onshore oil and gas industry facilities (e.g., refineries), offshore platforms are relatively simple, involving physical separation rather than chemical transformation. Offshore platform risk stems primarily from compactness and remoteness. Compactness contributes to ignition risk following a hydrocarbon leak. Remoteness contributes to weather exposure and evacuation difficulty.

The major functions of ARMS are the following:

- Improve maintenance scheduling
- Monitor risk levels
- Detect abnormal situations in a timely manner
- Respond appropriately to abnormal situations.

ARMS extends existing deterministic platform safety systems whose last resort is an emergency shutdown (ESD). An ESD, which can be triggered automatically (e.g., by fire alarms) or manually, cuts flow from the reservoir, thereby reducing a hydrocarbon leak's consequences.

ARMS provides additional response layers between normal operations and an ESD (Table 1). Deterministic real-time expert system reasoning continuously monitors risk and responds to low-severity abnormal situations. Normative system reasoning diagnoses abnormal situations from control system information and responds to intermediate-severity abnormal situations.

Table 1: Risk Management Response Layers

| Operating Status | Response |
| --- | --- |
| Normal Operations | Standard Procedures |
| Low-Severity Abnormal | ARMS Deterministic Expert System |
| Intermediate-Severity Abnormal | ARMS Normative System |
| High-Severity Abnormal | Emergency Shutdown |

Maintenance is an ongoing activity in the harsh marine environment and often is the root cause of abnormal situations such as leaks and fires (e.g., simultaneous welding and gas valve repair in nearby locations). ARMS's maintenance scheduling capability assists with daily planning. Automation provides platform managers with regular benefits, thereby motivating them to keep ARMS's information current. System recommendations are therefore more reliable when an abnormal situation of intermediate severity threatens or occurs.

### 3.2 GAS COMPRESSOR MODULE EXAMPLE

The greatest risk for offshore platforms is a hydrocarbon leak that ignites. The most difficult fire risk is a gas leak, which may be invisible and odorless. Two gas leak types are considered: progressive leaks start small and grow, catastrophic leaks are complete breaks. Gas detectors, process information (e.g., pressure detectors), and workers identify leaks. Leak response involves source identification and subsequent repair with minimal production interruption. Various production shutdown levels balance safety and production.

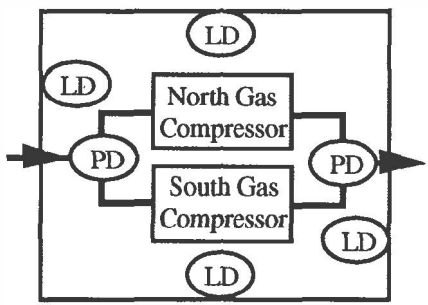

LD: Gas Level Detector

PD: Gas Pressure Detector

──── Piping

Figure 3: Gas Compressor Module Layout

A simplified gas compressor module illustrates the problem (Figure 3). Gas enters, flows through two parallel compressors, and exits. Pressure detectors (PD) monitor internal gas flow and level detectors (LD) monitor external released gas.

After a leak alert, the principal decision is the appropriate shutdown level given observations (automatic and human) and dynamic leak evolution. Information gathering refines the principal decision recommendation.

## 4 NORMATIVE SYSTEM OVERVIEW

An influence diagram (ID) template (Figure 4) integrates diagnosis, dynamic evolution, decision making and information gathering. The template formalizes the engineering risk management tradeoffs implicit in Figure 1.

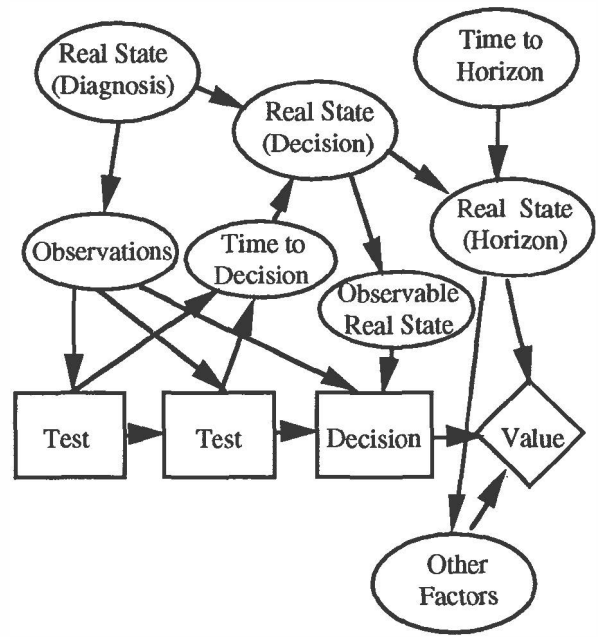

Figure 4: Risk Management Influence Diagram Template

Diagnosis infers the real physical state (e.g., gas leak volume and ignition status) from observations. Given a probabilistic diagnosis at a particular time, dynamic evolution generates a real-state distribution at decision time and at a finite time horizon. The real state may be partially observable (e.g., leak ignition). Decision making determines the appropriate action given the real state and other pertinent (e.g., financial) factors. Information gathering identifies tests that refine the decision-making recommendation.

## 5 NORMATIVE SYSTEM ACTIVITIES

Although the problem can be formulated and evaluated in a single ID, natural independence assertions (discussed in subsections) enable each analysis activity to be treated with a tailored approach (Table 2).

Table 2: Activity Models and Evaluation

| Activity | Model | Evaluation |
|---|---|---|
| Diagnosis | Belief Network | Message-Passing Inference |
| Dynamic Evolution | Dynamic Model | Dynamic Iteration |
| Decision Making | Influence Diagram | Arrow-Reversal Inference |
| Information Gathering | Decision Tree | Decision Tree Rollback |

### 5.1 DIAGNOSIS

Given observations, probabilistic inference in a belief network (BN) generates a distribution over the real state (Figure 5).

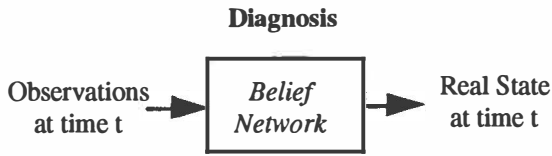

Figure 5: Diagnosis Function

Given that a compressor module gas leak has not ignited, the leak volume is not immediately apparent, and hence presents a diagnosis challenge. (If the leak ignites, diagnosis on a compact offshore platform is irrelevant.) A simple BN infers the leak state from observations (Figure 6). The Detailed Leak State node contains a distribution over leak types (e.g., no leak; progressive or catastrophic leak in a compressor or piping section). Given each leak type, the observation nodes characterize automatic sensor and manual inspection reliability. Component-specific leak states are aggregated into three Leak States—none, progressive, and catastrophic—for decision-making purposes, thereby reducing problem size.

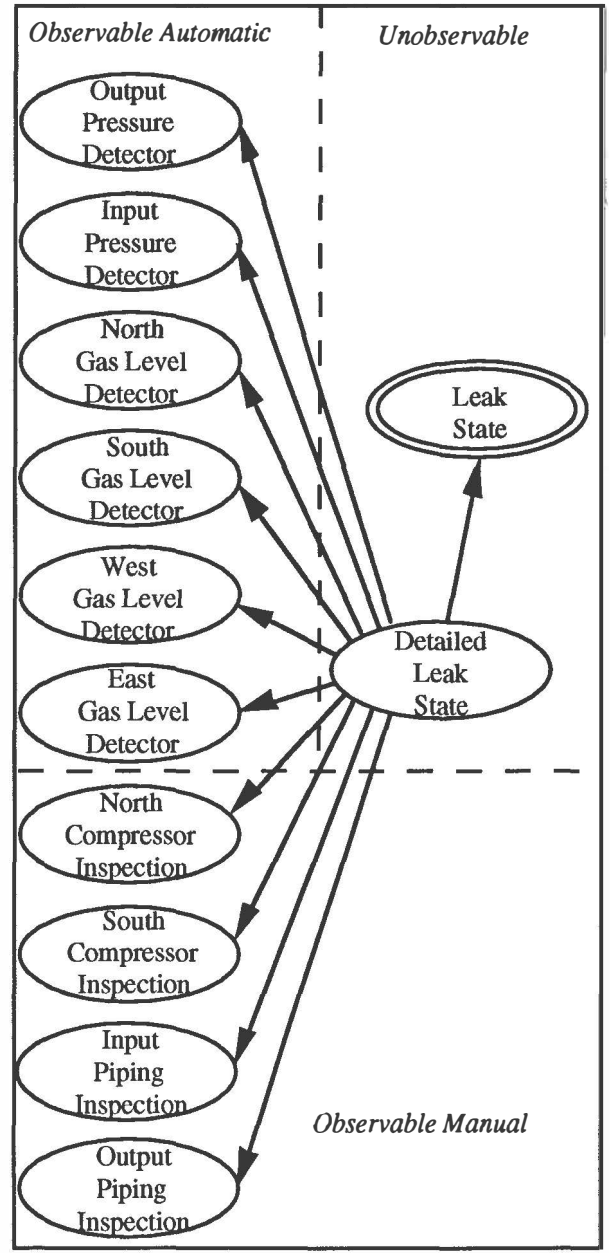

Figure 6: Diagnosis Belief Network

This BN represents conditional independence of observations given leak state. More complex dependence structure is possible. The use of simple activity models is deliberate to highlight the interaction among activities.

The diagnosis activity uses the message-passing inference method (Pearl 1988, Lauritzen and Spiegelhalter 1988). Effective methods exist to compose and construct diagnostic models (Heckerman 1990).

## 5.2 DYNAMIC EVOLUTION

Given a probability distribution over the real state at a particular time and a time delay, a dynamic model generates a state distribution at a later time (Figure 7).

**Dynamic Evolution**

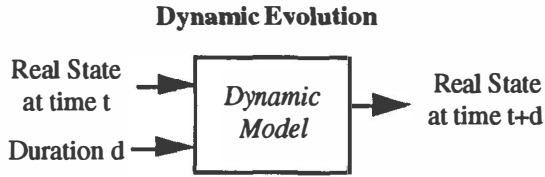

Figure 7: Dynamic Evolution Function

For the gas compressor module, a simple Markov model represents the leak's dynamic evolution (Figure 8). In each time interval, a progressive leak may become catastrophic, and either leak type may ignite. Increasing shutdown level decreases transition probabilities (p,q,r,s).

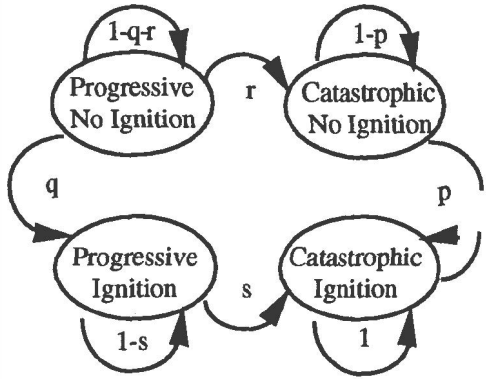

Figure 8: Dynamic Evolution Markov Model

The model's simplicity and separability are due to conditional independence of future leak state given prior leak state and a time period. More complicated dynamic models (Howard 1971) would increase the system's accuracy. Accuracy must be balanced against increased computational requirements.

## 5.3 DECISION MAKING

Given alternatives, information (including diagnosis results), and decision maker preferences, ID evaluation generates a recommendation (Figure 9).

**Decision Making**

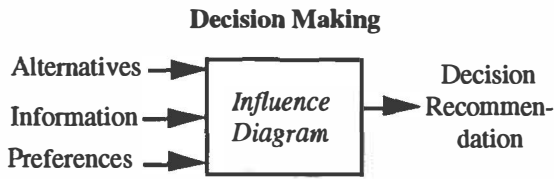

Figure 9: Decision-Making Function

For the gas compressor module leak, the principal decision is the shutdown level (Figure 10). The more extensive the shutdown the lower the ignition risk (with its personnel, equipment, future production, and environmental losses) and the greater the certain current production loss. Safety versus production is the basic tradeoff.

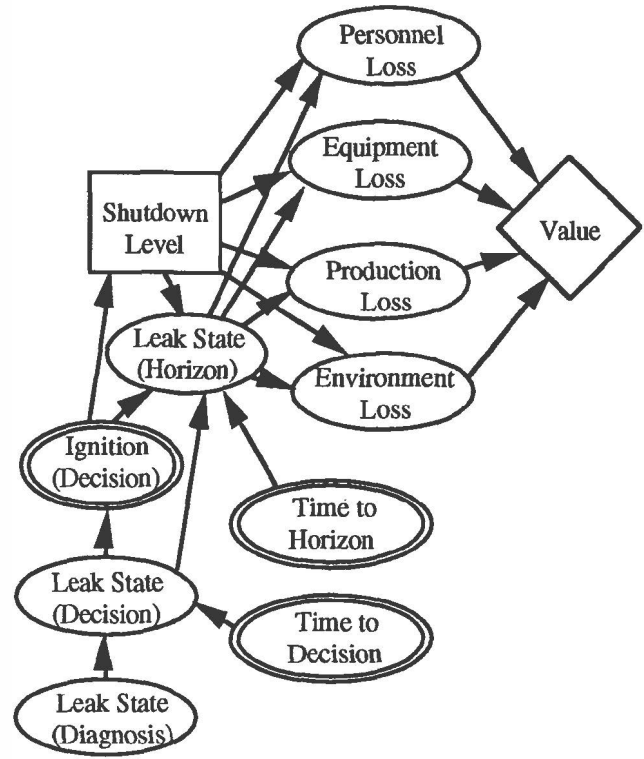

Figure 10: Decision-Making Influence Diagram

The principal decision is the ERMS's core (Figure 4). The diagnosis function provides an initial Leak State distribution that the dynamic evolution function projects forward to the time of the decision. The Leak State at decision time is partially observable in that ignition is evident. The Shutdown Level decision determines the transition probabilities for each alternative (e.g., slower leak evolution for more extensive shutdown). The Leak State at a finite horizon is conditional on no ignition at decision time and is projected forward by the dynamic evolution function. The decision horizon specifies how far into the future the analysis extends.

The appropriate horizon depends on the situation (e.g., the leak's severity or the availability of onshore assistance). A reasonable approach is to first consider a short horizon. If analysis recommends full shutdown, then act immediately. If not, then consider longer time horizons.

The decision-making activity uses the arrow-reversal ID solution method (Olmsted 1983, Shachter 1986). The dynamic evolution activity is a dynamic probabilistic function within the ID. Effective methods exist to design and construct decision models (Holtzman 1989, Regan 1993).

## 5.4 INFORMATION GATHERING

The information-gathering activity uses each previous activity and a test alternative knowledge base (KB) to generate an information-gathering plan with principal decision recommendations that are contingent on test results (Figure 11). A decision tree efficiently captures sequential information-gathering asymmetries.

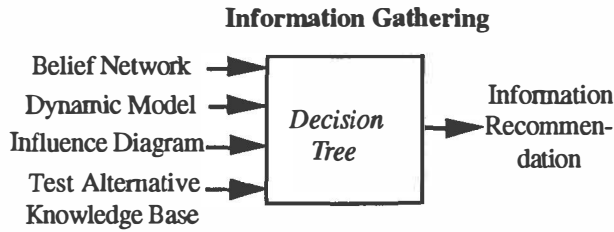

Figure 11: Information-Gathering Function

Other researchers have considered myopic (single test) and approximate nonmyopic (multiple tests) information gathering (Heckerman et al. 1990, Heckerman et al. 1991, Horvitz and Rutledge 1991). My objective is to manage the combinatorial complexity of exact nonmyopic information gathering by automating incremental problem formulation and evaluation. These steps correspond to reformulation and reevaluation in the broader decision analysis context (Section 2). These broader terms are used here.

Each path in the information-gathering decision tree specifies a test and observation sequence. The diagnosis function uses observations along a given path to generate a real-state distribution. The dynamic evolution function converts this distribution into an updated real-state distribution at the time of the principal decision. Given ignition prior to the decision, the decision-making function evaluates the consequences. Given no ignition, the dynamic evolution function projects the real state to the decision horizon and the decision-making function generates a principal decision recommendation.

The sequential information-gathering algorithm can reformulate and then reevaluate or it can interleave the tasks. Reformulation followed by reevaluation is simplest. The process starts with framing constraints (e.g., limits on test sequence length, time, or cost). Eligibility criteria in the test alternative KB identify test sequences that are consistent with the framing constraints. The diagnosis and dynamic evolution functions provide conditional probability distributions for test observations along each decision tree path. The decision-making function generates information-gathering and principal decision recommendations.

Interleaving reformulation and reevaluation is more complex but provides useful intermediate results. The process starts with a single information alternative (i.e., the myopic approach). After reevaluation, the algorithm identifies a path to expand, thereby transforming nonmyopic information gathering into a search task. The process expands the chosen path one step and then iterates.

Heuristics guide selection of the path to expand. For example, the path with the highest expected value could be chosen for expansion. In practice, a probabilistic strategy that distinguishes between uncertainties and decisions is more effective (Regan 1993).

Computational and judgmental features of the test alternative KB are important. First, test alternative screening reduces the search tree's breadth, thereby pushing back the computational frontier of this inherently combinatorial task. Second, experts should be better able to specify general test alternative eligibility criteria than to specify asymmetric test sequences that are appropriate across many decision instances. Eligibility criteria may be characterized in terms of the real state (e.g., high catastrophic leak probability). If so, the diagnosis and dynamic functions are evaluated for each path prior to screening. The search-pruning benefits must be balanced against the additional computation required.

The incremental construction of the information-gathering decision tree is a flexible algorithm with steady progress toward the complete solution (Horvitz 1991). The best solution so far is available at any given time. This approach is more flexible than a traditional tree rollback which requires formulating and solving an entire tree of a predefined depth before any results are available.

## 5.5 DISCUSSION

Separating the analysis activities allows tailored algorithms to be applied. Recent research that extends message-passing inference to decision problems (Shachter and Peot 1992) allows efficient diagnosis and decision making with a single algorithm. Similarly, asymmetric ID modeling (Smith et al. 1993) improves information-gathering efficiency.

Even in an ID environment that incorporates recent advances, separating activities has two major benefits: problem formulation decomposition and model economy. Decomposing problem formulation highlights each activity's primary feature (Table 3).

Activity separation economizes by obviating the need to repeat the entire BN for each diagnosis scenario in a comprehensive ID. Instead, a single BN is evaluated under many scenarios and the other activities use as appropriate the resulting posterior observation and real-state distributions.

Table 3: Activity Features

| Activity | Primary Feature |
|---|---|
| Diagnosis | Information Quality |
| Dynamic Evolution | Time |
| Decision Making | Value |
| Information Gathering | Asymmetry |

## 6  RESEARCH ISSUES

Normative system scaleup to handle a problem of realistic complexity presents domain-dependent (e.g., modeling gas leak evolution) and general challenges. Most importantly, nonmyopic information gathering is exponentially complex and is highly dependent on the complexity of the diagnosis, dynamic evolution, and decision-making models. Application to static decision situations would reduce computational complexity.

Interactive IDS formulation guidance can be combined with the ERMS architecture. Starting with fixed templates that represent the links between diagnosis and decision making (e.g., the Leak State node in Figures 6 and 10), IDS techniques can guide the user in model refinement for a particular decision.

A good design is necessary but not sufficient for successful development of any product, including normative systems. Careful attention must be paid to the construction tools and procedures for transforming a design into a working software system. An efficient construction methodology and an associated set of software engineering tools are under development for rapid development of domain-specific intelligent decision applications from a domain-independent core (Regan 1993).

## 7  CONCLUSIONS

This paper brings together diagnosis-focused and decision-focused normative systems in a broad framework. Removing the dynamic function, simplifying the principal decision model, and restricting information gathering to the myopic case yields current normative diagnostic systems. Removing the dynamic and diagnostic functions, and introducing interactive principal decision model formulation yields current intelligent decision systems. The appropriate balance among the four major activities depends on the particular application domain.

The principal conclusions are as follows:

- A normative system formalizes the dynamic tradeoffs characteristic of engineering risk management.
- Diagnosis, dynamic evolution, decision making, and information gathering are modeled and evaluated separately in a coherent normative system design.
- Automated incremental problem formulation and evaluation manage the combinatorial complexity nonmyopic information gathering.
- The normative system is embedded in a real-time expert system to appropriately balance deterministic and probabilistic reasoning.
- This approach can be applied to other domains (e.g., medicine, business).

### Acknowledgements

I thank Dr. Elisabeth Paté-Cornell, Dr. Abder Seridji, and Mr. William Gale for their contribution to the ideas developed in this paper. Grants from the National Science Foundation (SEF-9110462) and Bureau Veritas, Paris supported this work. I also thank my colleagues at Strategic Decisions Group who permitted me to use their software platform to refine and implement my ideas.

### References

Besse, P., et al., (1992). The Advanced Risk Management System Project: A Resource-Constrained Normative Decision System. *Proceedings of the THERMIE Conference on Oil and Gas Technology in a Wider Europe.* Berlin, Germany.

The Hon. Lord Cullen (1990). *The Public Inquiry into the Piper Alpha Disaster, Volumes 1 and 2 (Report to Parliament by the Secretary of State for Energy by Command of Her Majesty).* London, England.

Heckerman, D.E. (1990). *Probabilistic Similarity Networks.* MIT Press, Cambridge, MA.

Heckerman, D., E.J. Horvitz, and B.N. Nathwani (1990). Toward normative expert systems: The Pathfinder project. Technical Report KSL-90-08, Medical Computer Science Group, Section on Medical Informatics, Stanford University, Stanford, CA.

Heckerman, D.E., E.J. Horvitz, and B. Middleton (1991). An approximate nonmyopic computation for value of information. In D'Ambrosio et al., eds., *Uncertainty in Artificial Intelligence: Proceedings of the Seventh Conference,* pp. 135-141. Morgan Kaufmann, San Mateo, CA.

Henley, E.J., and H. Kumamoto (1981). *Reliability Engineering and Risk Assessment.* Prentice-Hall, Englewood Cliffs, NJ.

Holtzman, S. (1989). *Intelligent Decision Systems.* Addison-Wesley, Reading, MA.

Horvitz, E.J. (1991). *Action Under Bounded Resources.* PhD thesis, Section on Medical Informatics, Stanford University, Stanford, CA.

Horvitz, E.J., and D. Rutledge (1991). Time-dependent utility and action under uncertainty. In D'Ambrosio et al., eds., *Uncertainty in Artificial Intelligence: Proceedings of the Seventh Conference*, pp. 151-158. Morgan Kaufmann, San Mateo, CA.

Howard, R.A. (1966). Decision analysis: Applied decision theory. In D.B. Hertz and J. Melese, eds., *Proceedings of the Fourth International Conference on Operational Research*, pp. 55-77. Also in Howard, R.A., and J.E. Matheson, eds. (1981), *Readings on the Principles and Applications of Decision Analysis* 1: 97-113. Strategic Decisions Group, Menlo Park, CA.

Howard, R.A. (1971). *Dynamic Probabilistic Models, Volumes 1 and 2*. John Wiley and Sons, New York, NY.

Howard, R.A., and J.E. Matheson, eds. (1981). *Readings on the Principles and Applications of Decision Analysis*. Strategic Decisions Group, Menlo Park, CA.

Lauritzen, S.L., and D.J. Spiegelhalter (1988). Local computations with probabilities on graphical structures and their application to expert systems. *Journal of the Royal Statistical Society* 50: 157-224.

Olmsted, S.M. (1983). *On Representing and Solving Decision Problems*. PhD thesis, Department of Engineering-Economic Systems, Stanford University, Stanford, CA.

Paté-Cornell, M.E. (1990). Organizational aspects of engineering system safety: the case of offshore platforms. Science 250: 1210-1217.

Pearl, J. (1988). *Probabilistic Reasoning in Intelligent Systems: Networks of Plausible Inference*. Morgan Kaufmann, Menlo Park, CA.

Regan, P.J. (1993). *Design and Construction of Normative Risk Management and Decision Systems*. PhD thesis, Department of Engineering-Economic Systems, Stanford University, Stanford, CA.

Regan, P.J. and S. Holtzman (1992). R&D Analyst: An interactive approach to normative decision system model construction. In Dubois, D. et al., eds., *Uncertainty in Artificial Intelligence: Proceedings of the Eighth Conference*, pp. 259-267. Morgan Kaufmann, San Mateo, CA.

Shachter, R.D. (1986). Evaluating influence diagrams. *Operations Research* 34: 871-882.

Shachter, R.D. and M.A. Peot (1992). Decision making using probabilistic inference methods. In Dubois, D. et al., eds., *Uncertainty in Artificial Intelligence: Proceedings of the Eighth Conference*, pp. 276-283. Morgan Kaufmann, San Mateo, CA.

Smith, J.E., S. Holtzman, and J.E. Matheson (1993). Structuring conditional relationships in influence diagrams. *Operations Research* 41(2).

US Nuclear Regulatory Commission (1975). *Reactor Safety Study WASH-1400 (NUREG/74/104)*. National Academy Press, Washington, DC.